\documentclass{article}
\usepackage{graphicx}
\usepackage{subfig}
\usepackage{algorithmic}
\usepackage{xspace}
\usepackage{caption}
\usepackage{siunitx}
\usepackage{cite}
\usepackage{amsmath,amssymb,amsfonts}
\usepackage{multirow}
\usepackage{xcolor}
\usepackage{microtype}
\date{}

\begin{document}
\title{Residual Deep Convolutional Neural Network for EEG Signal Classification in Epilepsy\thanks{This work is supported by Chinese Scholarship Council (CSC), Center for Personalized Translational Epilepsy Research (CePTER), and the Johanna Quandt Foundation. Email: (elu, triesch)@fias.uni-frankfurt.de}}

\author{Diyuan Lu$^{1, 2, 3}$, Jochen Triesch$^{1, 2, 3}$  \\
	\small $^{1}$Frankfurt Institute for Advanced Studies (FIAS), Frankfurt am Main, 60438, Germany \\
	\small $^{2}$Goethe University Frankfurt, 60438 Frankfurt am Main, Germany \\
	\small $^{3}$Center for Personalized Translational Epilepsy Research (CePTER), Frankfurt am Main, Germany \\
}

\maketitle              
\begin{abstract}
Epilepsy is the fourth most common neurological disorder, affecting about \SI{1}{\percent} of the population at all ages.  As many as \SI{60}{\percent} of people with epilepsy experience focal seizures which originate in a certain brain area and are limited to part of one cerebral hemisphere. In focal epilepsy patients, a precise surgical removal of the seizure onset zone can lead to effective seizure control or even a seizure-free outcome. Thus, correct identification of the seizure onset zone is essential. For clinical evaluation purposes, electroencephalography (EEG) recordings are commonly used. However, their interpretation is usually done manually by physicians and is time-consuming and error-prone. In this work, we propose an automated epileptic signal classification method based on modern deep learning methods. In contrast to previous approaches, the network is trained directly on the EEG recordings, avoiding hand-crafted feature extraction and selection procedures. This exploits the ability of deep neural networks to detect and extract relevant features automatically, that may be too complex or subtle to be noticed by humans. The proposed network structure is based on a convolutional neural network with residual connections. We demonstrate that our network produces state-of-the-art performance on two benchmark datasets, a dataset from Bonn University and the Bern-Barcelona dataset. We conclude that modern deep learning approaches can reach state-of-the-art performance on epileptic EEG classification and automated seizure onset zone identification tasks when trained on raw EEG data. This suggests that such approaches have potential for improving clinical practice.
\end{abstract}


\section{Introduction}

Epilepsy is a chronic neurological disorder of the brain characterized by recurrent seizures. According to the World Health Organization (WHO), an estimated 2.4 million people are diagnosed with epilepsy each year worldwide. People  with  epilepsy  have  increased  morbidity and mortality, decreased social participation and frequently suffer from psychiatric and other co-morbidities and stigma. Sudden  unexpected  death  in  epilepsy  (SUDEP)  is  a  fatal complication of epilepsy, one of the most frequent causes of death in younger epilepsy patients. 

Diagnosing the presence and type of epilepsy is pivotal for successful treatment. In particular, electroencephalography (EEG) recordings provide a useful diagnostic tool. Trained neurologists interpret the EEG recordings through visual inspection for seizure types and ictal, inter-ictal phases etc. This approach is time consuming, subjective, and prone to errors. Therefore, automatic interpretation and classification of EEG signals in epilepsy patients is highly desirable.

\begin{figure}[t]
	\centering
	\includegraphics[width=0.8\textwidth]{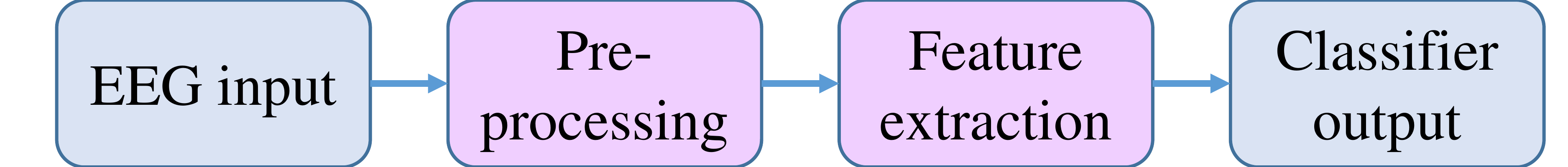}
	\caption{A typical EEG classification framework}
	\label{typical_work_flow}
\end{figure}

A typical system for EEG classification is shown in Fig.~\ref{typical_work_flow}. The achievable quality of the final classification is limited by the quality of preprocessing and feature extraction. Often, the features to be extracted are selected in an {\em ad hoc} fashion without any guarantee that these features are optimal for the task at hand and potentially constrain the performance.
Recently, however, fueled by advances in machine learning, systems have achieved impressive results by skipping the step of extracting pre-defined features. Instead, these systems are trained in an end-to-end fashion and {\em learn} to extract relevant features directly from the preprocessed data. In many domains, such machine learning systems are now outperforming humans on difficult pattern recognition tasks. In particular, so-called deep convolutional neural networks learn to extract local ``low-level'' features from the raw input and then progressively extract more global ``high-level'' features in successive processing layers.

Here, our goal is to test the utility of modern deep learning methods for the problem of EEG classification in the context of epilepsy and to compare them to the state-of-the-art. Specifically, we make the following contributions. We:
\begin{itemize}
	\item propose a neural network architecture using so-called residual connections to classify raw preprocessed EEG data without prior feature extraction,
	\item validate our framework on two published datasets (Bonn dataset \cite{andrzejak2001indications} and Bern-Barcelona dataset \cite{andrzejak2012nonrandomness}) achieving state-of-the-art performance.
	\item investigate how the network learns to solve these tasks by inspecting the response properties of units in the network.
\end{itemize}

The remainder of the paper is organized as follows. Section~\ref{database} describes the datasets that are used in this study. We describe our methods in Section~\ref{methods} and report the results of our experiments in Section~\ref{results} comparing them to published results. Section~\ref{discussion} gives a discussion.

\section{Datasets}\label{database}


\subsection{Bonn EEG Dataset}
This dataset was collected by Andrzejak
et al.\ \cite{andrzejak2001indications} at Bonn University, Germany\footnotemark. 
\footnotetext{\cite{andrzejak2001indications} (http://epilepsy.uni-freiburg.de/database)}
It consists of two sets of surface single-channel EEG segments from healthy volunteers with and without eyes open (sets A and B) and three sets of intra-cranial EEG recordings from epilepsy patients during the seizure-free period from within and outside the seizure onset zone (sets D and C) as well as during seizures (set E). Each set consists of 100 segments of 23.6 second-long signal recorded under a sampling rate of \SI{173.61}{\hertz}. A band-pass filter with frequency range 0.53 -- \SI{40}{\hertz} is applied to all the signals. In this study, we group signals from sets A and B into a class labeled {\em ``healthy''}, and those from sets C and D into a class {\em ``unhealthy''}, and those from set E into a class {\em ``seizure''} as suggested in \cite{acharya2017deep}.

\subsection{Bern-Barcelona Dataset}
The EEG segments from the Bern-Barcelona dataset were recorded with intra-cranial electrodes from five patients with focal epilepsy, and collected by Andrzejak et al.\ \cite{andrzejak2012nonrandomness}. A more challenging task is addressed in this dataset, {\em i.e.}, distinguishing signals recorded either from the epileptic focus or not, given only recordings during resting period. The clinical purpose of these intra-cranial recordings is to identify the brain areas that are involved in seizure initiation without an actual seizure being generated. Since quite often the occurrence of seizures is very low and seizures by themselves could cause brain damage. Therefore, correct identification could shorten the diagnosis period and minimize the damage caused by seizures. An additional advantage of this dataset is that numerous studies have already been conducted with it \cite{chua2009automatic, acharya2011automatic, juarez2015epilepsy, sharma2015integrated, acharya2018deep}.

The segments from this database are divided into two classes based on where they are recorded, namely from within the epileptic focus (focal) or from outside (non-focal). Each class contains 3750 segments with a duration of \SI{20}{\second} under a sampling rate of \SI{512}{\hertz} rendering 10240 data points per segment. Each segment contains data from two channels.

\section{Methods}\label{methods}
 In recent years, image classification and object recognition methods based on convolutional neural networks (CNNs) have achieved impressive results \cite{rawat2017deep}. The fundamental building block of a CNN is the Conv-layer. A kernel of fixed length (kernel size) slides over the input with a certain stride while performing a convolution operation on the input. This operation results in a feature map for each kernel. The number of channels represents the number of such kernels deployed to extract local features. Inspired by the nonlinear behavior of nerve cells, nonlinear activation functions enable the networks to learn complex, nonlinear mappings between the input and the output. To obtain invariance against shifted inputs (spatial shifts for images, temporal shifts for time series), max-pooling layers are added to the network. In order to prevent the network from over-fitting the training data and achieve better generalization capability, batch-normalization \cite{ioffe2015batch} and drop-out \cite{JMLR:v15:srivastava14a} are popular methods. Typically, a number of fully connected layers are used after the convolutional layers. The final layer typically outputs a probability distribution over all possible categories for the given input.

\subsection{Proposed Network Structure}
\begin{figure}[tb]
	\centering
	\includegraphics[angle=90, scale=0.55]{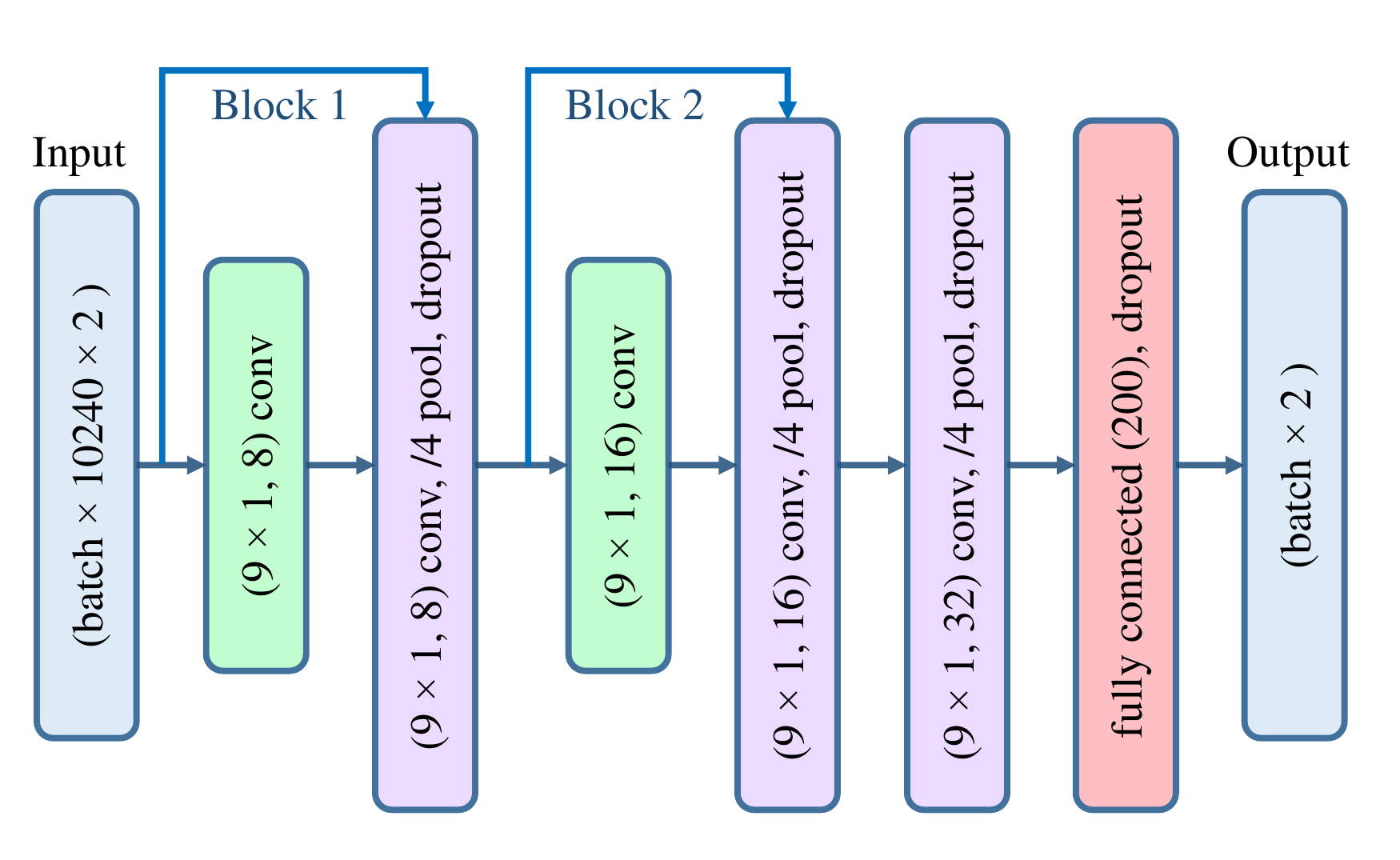}
	\caption{The structure of the proposed network. The shapes of the input, output, and convolutional kernels of the network are given in parentheses. }
	\label{proposed}
\end{figure}

We propose a deep convolutional neural network architecture with so-called residual blocks \cite{he2016deep} for EEG data classification (Fig.~\ref{proposed}). There are two residual blocks in the network. Each block consists of two Conv-layers, one max-pooling, and one drop-out layer at the end.  In the diagram, $(9 \times 1, 8)$ represent the kernel size is in shape $9\times 1$, and the number of kernels used in the layer is 8. Notation $/4$ in the pooling layer means that the down sampling factor is four. We perform batch normalization after every Conv-layer before pooling layer. For clarity, we omit the batch normalization layer in the structure schematic. The input shape is given for the Bern-Barcelona dataset. We keep the network hyperparameters fixed for both experiments.

The EEG signals are time series recorded from one or more electrodes. The number of electrodes can be viewed as data width and length of the recording is the data height. We perform 1D convolution along the time axis (height) of the EEG data. The dimension reduction in the pooling layer is done by a pooling window of 4 with a stride of 4 along the time axis, which worked best among different configurations we have tried. 
We use the Adam optimizer with a mini-batch size of 20. Preliminary experiments showed that larger batch-sizes do not improve the testing accuracy. We adopt a learning rate decay strategy, starting with a learning rate of 0.01 and multiplying it by 0.1 three times at epochs 10, 30, and 50. The weight initialization used in this experiment is the default tensorflow Xavier weight initializer. The LReLU \cite{maas2013rectifier} is applied across all layers after batch normalization except for the last FC layer where softmax activation is used to output a probability distribution among possible classes.

In order to increase the amount of training data, we applied a random crop strategy on the original data on both datasets. Crop-size is the crucial parameter. Here we pick 3800 samples for the Bonn dataset and 9800 for the Bern-Barcelona set, since it theoretically increases the number of signals by a factor of (original length - crop-size) and prevents the model from over-fitting. EEG recordings are all normalized to zero mean and a standard deviation of one using the Z-score normalization method before being fed into the network. The proposed method is implemented in Python with TensorFlow and runs on a desktop computer with Intel Core i7 870 CPU @ \SI{2.93}{\giga\hertz} $\times$ 8 with 4.0 GB RAM. 

\section{Results}\label{results}


\subsection{Experiment on the Bonn Dataset}
On the Bonn dataset, we performed a three-class classification task. We grouped set A and set B together as healthy recordings with label 0. We group set C and set D to form unhealthy data with label 1. Set E is the class of seizure data with label 2. From each set, we randomly take \SI{60}{\percent} for training (300 segments),  \SI{20}{\percent}  for testing (100 segments) and the remaining \SI{20}{\percent} for validation (100 segments). In every training batch, we choose the crop-size as 3800 data points. This number works relatively better among different crop-sizes we have tried. A comparison of classification accuracy of several studies is depicted in Table~\ref{comparison_Bonn}. Our method outperforms previous published methods.

\begin{table}[t]
	\centering
	\caption{Comparison of different approaches on the Bonn dataset. Highest score is highlighted in bold face.}
	\captionsetup{justification=centering}
	\begin{tabular}{|l|l|l|}
		\hline
		\textbf{Reference}	 & \textbf{ Method} & \textbf{Accuracy ( \SI{}{\percent})}  \\
		Chua et al. (2009) \cite{chua2009automatic}  & HOS + GMM &93.1 \\
		Acharya et al.(2011) \cite{acharya2011automatic} & DWT + SVM & 96.3   \\
		Juarez-Guerr et al. (2013) \cite{juarez2015epilepsy} & DWT + FNN  & 93.23 \\
		R. Sharma et al. (2015) \cite{sharma2015integrated}  & EMD  Least square - SVM & 98.67 \\
		Acharya et al. (2018) \cite{acharya2018deep} & CNN with raw EEG & 88.67  \\
		D Ahmedt-Aristizabal et al. (2018) \cite{ahmedt2018deep} & LSTM with raw EEG & 95.53\\
		\textbf{Proposed}   & \textbf{Res-CNN on raw EEG} & 	\textbf{99.0}\\
		\hline
	\end{tabular}
	\label{comparison_Bonn}
\end{table}

\begin{figure}[t]
	\centering
	\captionsetup[subfigure]{width=0.95\textwidth}
	\subfloat[]
	{
		\includegraphics[width=0.45\textwidth]{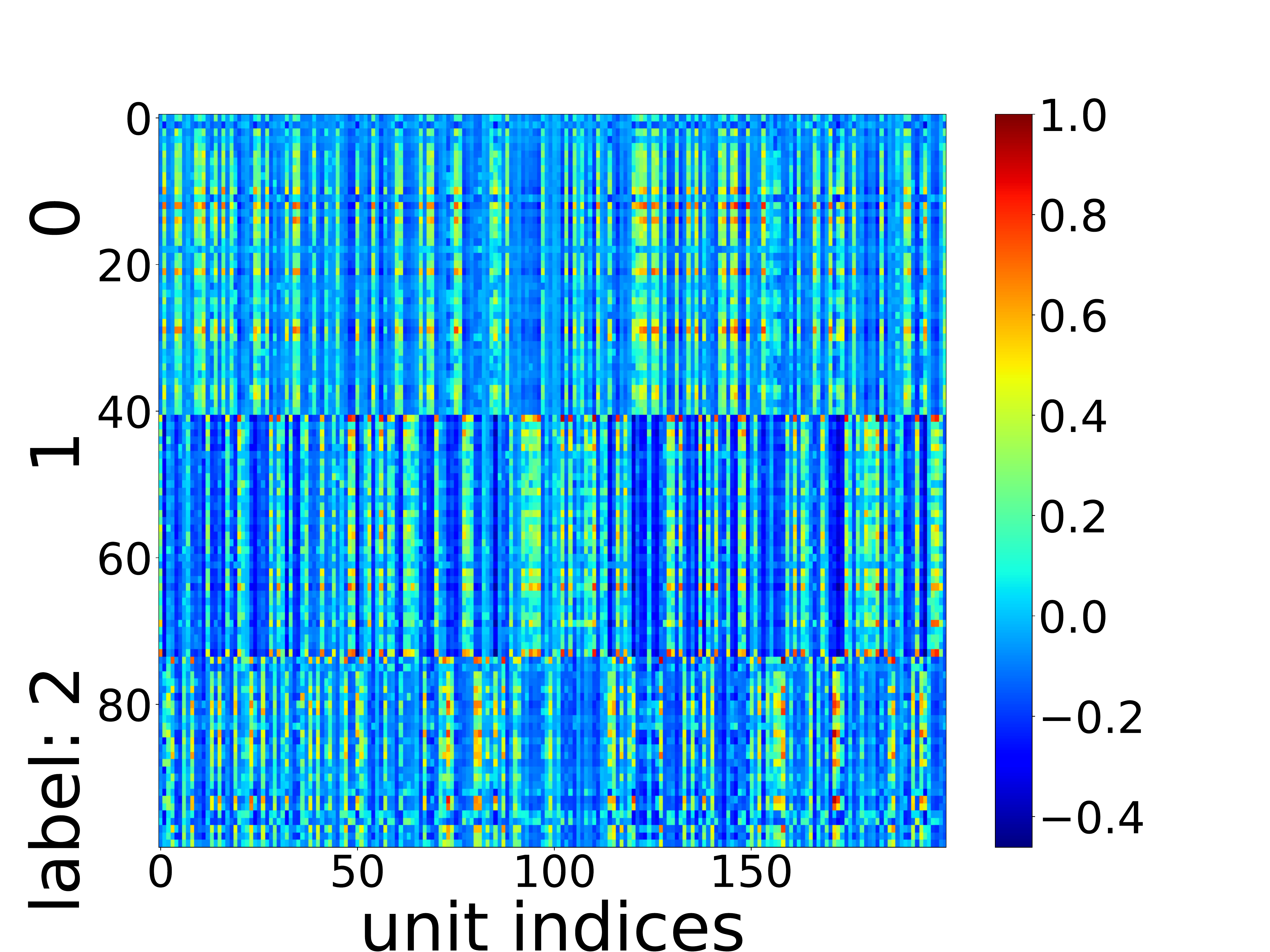}
		\label{Bonn-imshow}
	}
	\subfloat[]
	{
		\includegraphics[width=0.45\textwidth]{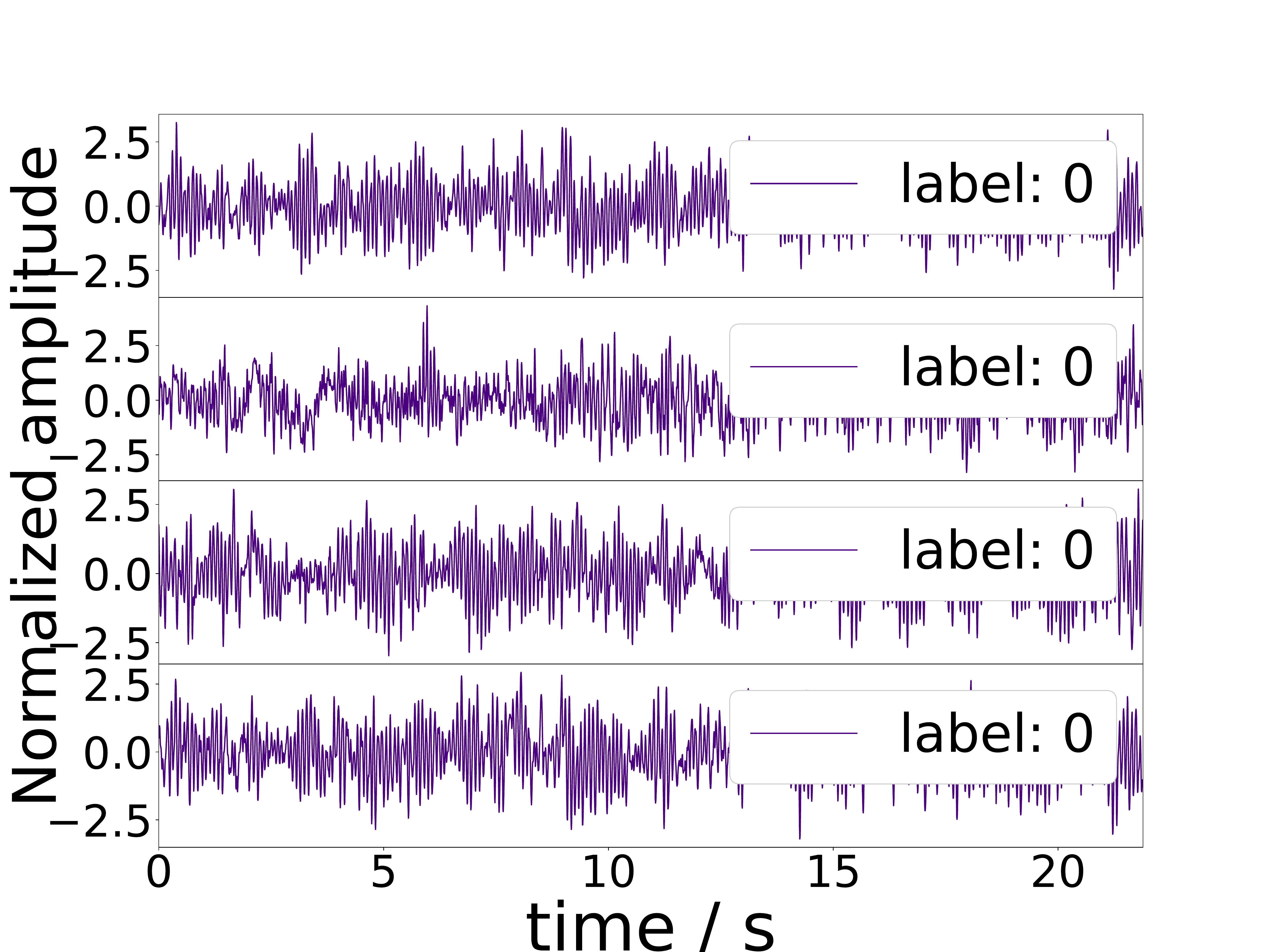}
		\label{Bonn-healthy}
	}\\
	\vspace{-4mm}
	\subfloat[]
	{
		\includegraphics[width=0.45\textwidth]{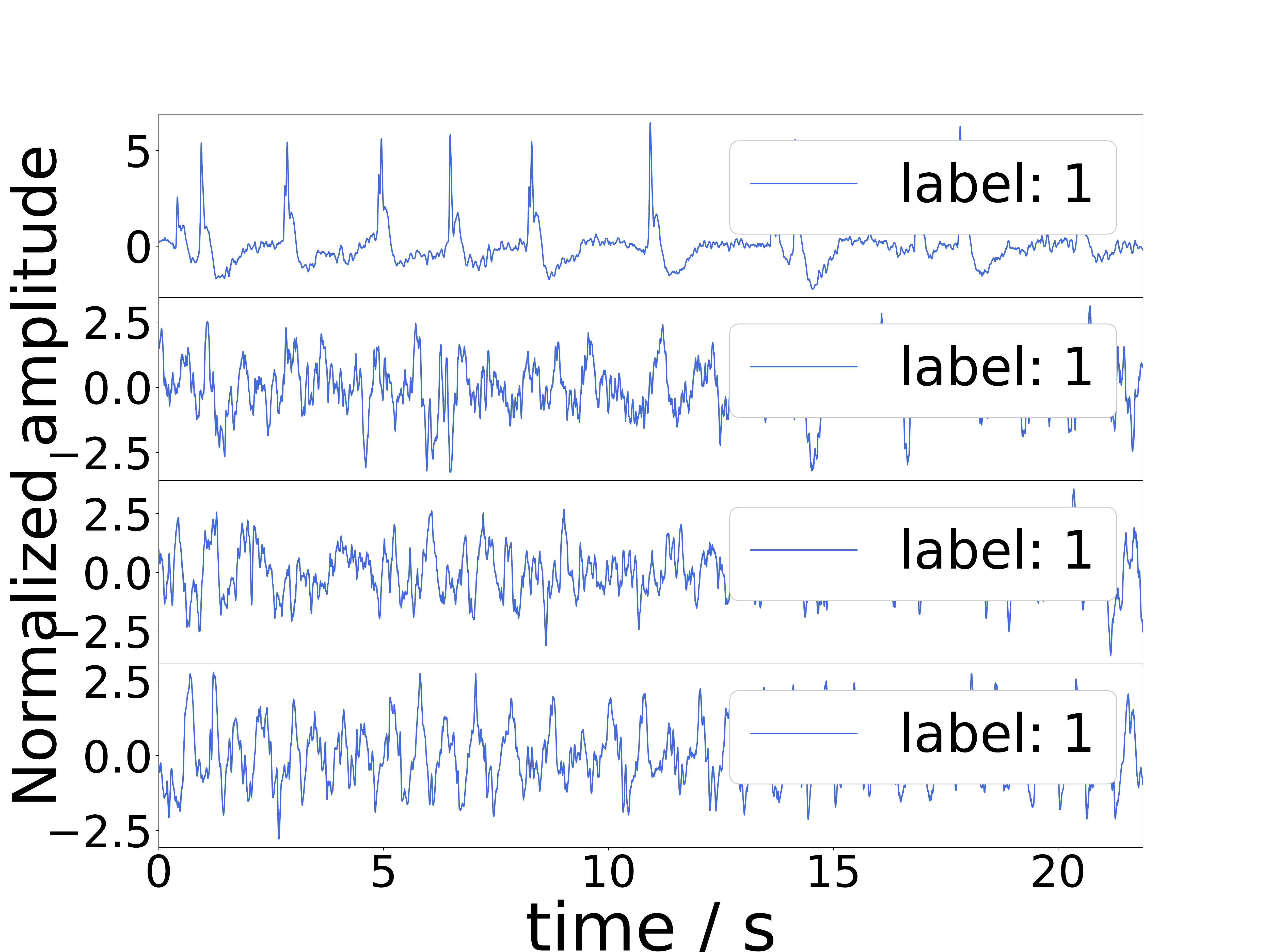}
		\label{Bonn-unhealthy}
		
	}	
	\subfloat[]
	{
		\includegraphics[width=0.45\textwidth]{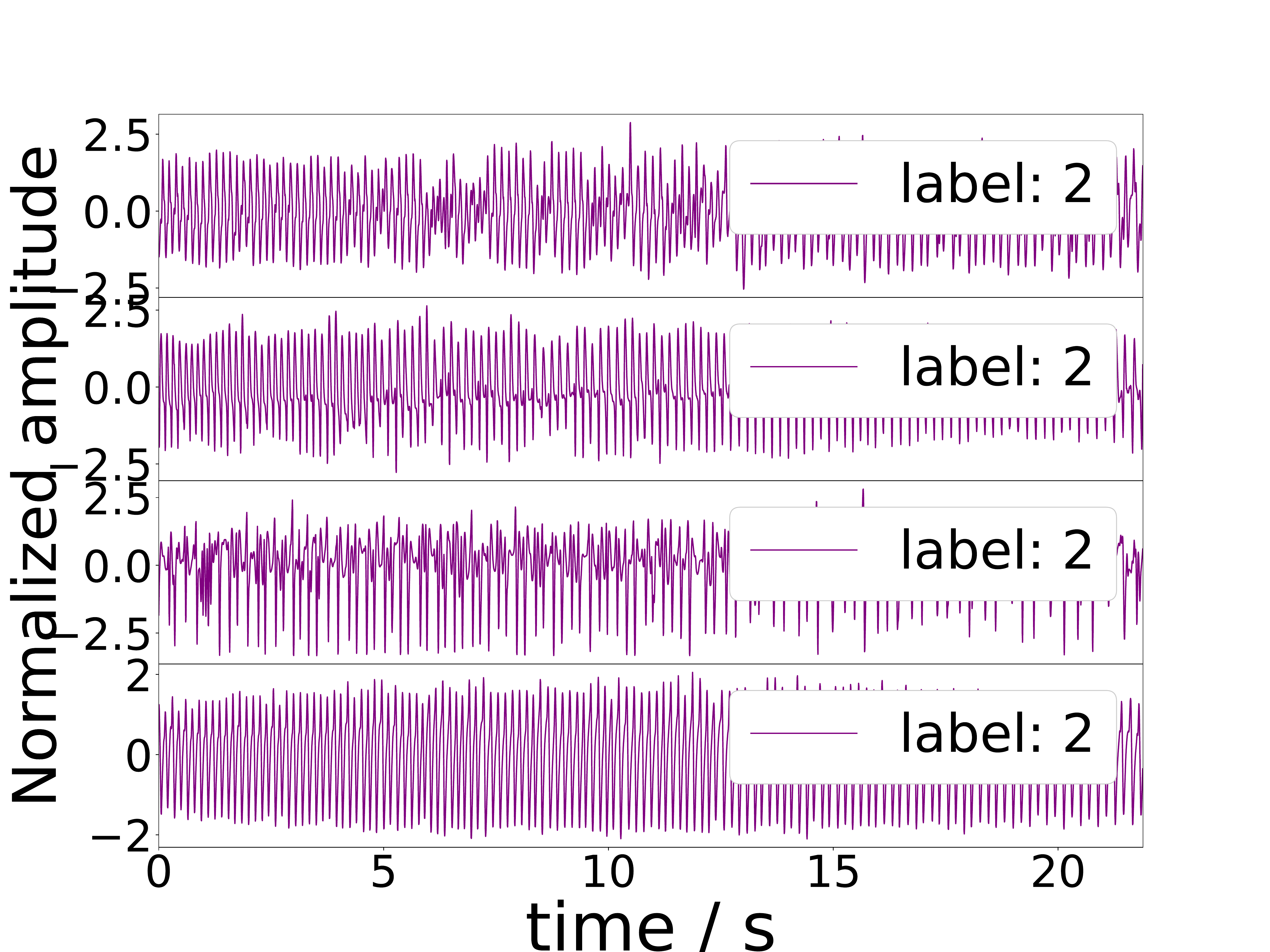}
		\label{Bonn-seizure}
	}
	\caption{Visualization of activity in the last FC layer on the Bonn dataset. $(a)$ Heatmap of activities generated by inputs belonging to the three different classes: 0 -- {\em ``healthy''}, 1 -- {\em ``unhealthy''}, 2 -- {\em ``seizure''} (grouped vertically). Activity patterns in response to signals from the same class are highly correlated but are different across classes. $(b)$ -- $(d)$ EEG inputs that maximize the activity of example units \#14, \#110, and \#172, respectively. Each unit responds selectivley to inputs from a single class.}
	\label{Bonn-activity}
\end{figure}



Previous approaches typically extracted hand-crafted features from the original data as input to the classifiers.
In \cite{chua2009automatic}, Chua et al.\ extracted higher order spectral (HOS) features as the input and employed a Gaussian mixture model (GMM) as the classifier. They achieved a classification accuracy of  \SI{93.1}{\percent}. Archarya et al.\ in \cite{acharya2011automatic} extracted features using the discrete wavelet transformation (DWT) and employed a support vector machine (SVM) as the classifier. This approach led to an accuracy of  \SI{96.3}{\percent}. In their later study, they tried extracting entropy as well as HOS features and applied a fuzzy Sugeno model \cite{takagi1985fuzzy} as a classifier. This approach achieved an accuracy of \SI{89.04}{\percent}.
Sharma et al.\ \cite{sharma2015integrated} employed empirical mode decomposition (EMD) to extract features and a least square SVM (LS-SVM) as the classifier. This method achieved an accuracy of  \SI{98.67}{\percent}. Numerous experiments have been conducted based on extracting hand-crafted features, but working directly with raw EEG data is challenging and much less work has been done. In \cite{acharya2017deep}, Acharya proposed a 13-layer convolutional neural network to perform the same task. They achieved an accuracy of \SI{88.}{\percent}.
In our work, by designing a more efficient CNN structure and leveraging the residual connection approach\cite{he2016deep}, we reach an accuracy of  \SI{99.0}{\percent}, sensitivity \SI{96.15}{\percent} (TP / TP + FN), and specificity of \SI{100.0}{\percent} (TN / TN + FP) with only around 30 training epochs from the raw data. Compared to \cite{acharya2018deep}, the accuracy is improved by \SI{10.33}{\percent}.


In Fig.~\ref{Bonn-activity}, we show the activity of each individual unit in the last FC layer evoked by the test signals. Fig.~\ref{Bonn-imshow} shows a heatmap of the activity which is grouped by the signal class along the y-axis. The unit activities are well divided into subgroups that are most responsive to a single class. We then pick three exemplary units and plot the top four EEG input signals that maximize their response. Unit \#14 in Fig.~\ref{Bonn-healthy} responds most strongly to signals from the {\em healthy} class and unit \#110 in Fig.~\ref{Bonn-unhealthy} is activated strongly by signals from the {\em unhealthy} class. Lastly, unit \#172 is most active in response to recordings from the {\em seizure} class, as shown in Fig.~\ref{Bonn-seizure}.

\subsection{Experiment on the Bern-Barcelona Dataset}
The Bern-Barcelona dataset addresses a more challenging task since all signals are from patients who have focal epilepsy and the signals are recorded in the inter-ictal phase. We randomly take 500 pairs of samples as the test set and another 250 pairs as the validation set and the remaining 3000 pairs as the training set. Tne network is trained to output 0 for non-focal signals and 1 for focal ones. The random cropping strategy described in last section is also applied here. The crop-size is set to 9800 which works better than other crop-sizes we have examined.

Numerous studies have been conducted on the Bern-Barcelona dataset, since it has more recording samples and a high clinical significance.  A comparison of classification accuracy on the Bern-Barcelona dataset is given in Table~\ref{comparison_Barcelona}. To the best of our knowledge, there is no other study applying a neural network directly on the raw EEG signals in this dataset. Therefore, we compare our results to conventional feature extraction approaches. 

\begin{table}[t]
	\caption{Comparison of selected different approaches on the Bern-Barcelona dataset in recent years. Highest score is highlighted in bold face.}\label{tab1}
	\begin{tabular}{|l|l|l|}
		\hline
		\textbf{Reference}	 & \textbf{ Method} & \textbf{Accuracy (\SI{}{\percent})}\\
		Sharma et al. (2014) \cite{sharma2014empirical}  &Empirical mode decomposition (EMD)  & 87 \\
		M. Sharma et al. (2015) \cite{sharma2015integrated}  & Discrete wavelet transform (DWT) & 84\\
		AB. Das et al. (2016) \cite{das2016discrimination}	& EMD-DWT & 89.04       \\
		N. Sriraam (2017) \cite{sriraam2017classification}	& 21 features with SVM & \textbf{92.15}\\
		Bhattacharyya et al. (2018) \cite{bhattacharyya2018novel} & EME-DWT + SVM (50 pairs) & 90.0 \\
		Proposed  & Residual CNN on raw EEG & 91.8\\
		\hline
	\end{tabular}\label{comparison_Barcelona}
\end{table}

Sharma et al.\ \cite{sharma2014empirical} extracted sample entropy and variance features from the intrinsic mode functions obtained from EMD. Their study was conducted on 50 pairs of focal and non-focal recordings. These features were then fed into a LS-SVM classifier and achieved an accuracy of \SI{87.0}{\percent}. In their later work \cite{sharma2015integrated}, two classes of EEG signals are decomposed into six discrete frequency bands applying a DWT and entropy features are computed for each band. Selected features are used in a LS-SVM classifier achieving an accuracy of \SI{84}{\percent}.

In another study \cite{das2016discrimination}, Das et al.\ performed EMD on the EEG signals followed by DWT to compute log energy entropy. An accuracy of \SI{89.4}{\percent} was achieved applying a k-NN classifier. Sriraam et al.\ \cite{sriraam2017classification} investigated 26 potential features extracted from focal and non-focal signals. They performed a Wilcoxon rank sum test to identify significant features and Tukey's range test was adopted for removing feature outliers. Finally, 21 features were selected and used for classification with an SVM yielding an accuracy of \SI{92.15}{\percent}. Bhattacharyya et al.\ \cite{bhattacharyya2018novel} applied empirical wavelet transform (EWT) technique to decompose the original signal into rhythms. And then followed by a LS-SVM classifier to get the final classification result. These studies all depend on extracting features such as entropies, DWT or EMD, etc. In contrast, the proposed method learns directly from the input data with minimal pre-processing and reaches a comparable accuracy of \SI{91.8}{\percent}. The sensitivity and specificity achieved by this approach is \SI{95.3}{\percent} and \SI{87.7}{\percent}, respectively. 




\section{Discussion}\label{discussion}
The main contribution of this work is the implementation of a deep CNN model with residual connections that achieves state-of-the-art classification of EEG signals in the context of epilepsy. We have demonstrated its performance for classifying healthy, unhealthy, and seizure states using the Bonn dataset and classifying focal and non-focal recordings using the Bern-Barcelona dataset. The proposed network architecture combining convolutional, batch-normalization, residual connections, and drop-out layers provides good convergence and has the highest performance among numerous network structures that we have examined. Further improvements may be possible with optimal hyper-parameter search and different network architectures that specialize in capturing long temporal dependency in the input signals.

In comparison to the Bonn dataset, we achieve a lower accuracy on the Bern-Barcelona dataset. We suspect several reasons for this. First, in the Bonn dataset one class contains only surface EEG data and the others only contain intracranial EEG data. These differences in recording technique may make the task quite easy. Second, the data in the Bern-Barcelona set are more complex. One EEG sample consists of 10240 data points and within one sample the signal is highly variant. Third, we used the same network structure and hyper-parameters as for the Bonn dataset and did not tune the network parameters to this particular task. Since the two datasets are collected with different sampling rates, special hyper-parameter tuning and structure modification may be needed in order to achieve higher performance on the Bern-Barcelona dataset.

Importantly, our method achieves state-of-the-art results with minimal data preprocessing and without extracting any hand-crafted features. This is in stark contrast to most previous work in this area, which has relied on manual or semi-automatic selection of particular pre-processing and feature extraction steps. We have avoided this altogether by learning directly from the raw signals. This approach is much more flexible, of course, but tends to require more training data and careful task-dependent hyper-parameter search. Indeed, the addition of more training data is likely to further improve our results. A large, well-curated, and publicly available repository for such data could greatly boost research in this area.

Finally, epilepsy can have many different causes. In the future, it might be interesting to try classify different cause of epilepsy in individual patients based on EEG measurements. If this could be achieved, it would open up new opportunities for individualized treatment.

\section*{Acknowledgment}

This work is supported by the Chinese Scholarship Council (CSC), the Center for Personalized Translational Epilepsy Research (CePTER), and the Johanna Quandt Foundation. We thank Felix Rosenow and Florence Kleberg for discussions and feedback on an earlier version of the manuscript and Valentin Neubert for early discussions on automated EEG analysis.

%
%
%
\bibliographystyle{splncs04}
\bibliography{reference}
%
%
%
%
%
\end{document}